\documentclass[10pt,twocolumn,letterpaper]{article}
\usepackage[pagenumbers]{cvpr}

\title{Cross-device Collaborative Test-time Adaptation with Zeroth-order Optimization and Model Merging}

\author{Yu Mitsuzumi$^{1}$\thanks{Corresponding Author: Yu Mitsuzumi (yu.mitsuzumi@ntt.com)},~~~Akisato Kimura$^1$,~~~Yasuhiro Fujiwara$^1$,~~~Hisashi Kashima$^2$ \\
${}^\text{1}$NTT, Inc.~~~~${}^\text{2}$Kyoto University}
\usepackage{amsmath}	
\usepackage{amssymb}
\usepackage{booktabs}
\usepackage{color, colortbl}
\usepackage{multirow}

\usepackage{algorithm}
\usepackage{algpseudocode}

\usepackage{mathtools}
\usepackage{siunitx} 
\newcommand{\nbf}[1]{{\noindent \textbf{#1}}}

\usepackage{xr-hyper}

\usepackage{hyperref}

\usepackage{orcidlink}

\usepackage[capitalize]{cleveref}
\crefname{section}{Sec.}{Secs.}
\crefname{table}{Table}{Tables}
\crefname{figure}{Fig.}{Figs.}

\begin{document}

\maketitle

\begin{abstract}
Test-time adaptation (TTA) mitigates domain shifts by using incoming test data to update a model on the fly. The majority of TTA methods require resource-intensive backpropagation (BP) for model updates, particularly demanding large memory sizes, which makes it \textit{infeasible} to deploy them on resource-limited devices (e.g., edge devices). To address this issue, we integrate two different techniques, zeroth-order optimization (ZOO) and model merging, under the recently established cross-device collaborative TTA (CDC-TTA) framework, where the system is composed of a mixture of resource-abundant and resource-limited devices, and the model information (e.g., model weights obtained on each device) is shared across the devices.
Our method is \textit{executable} on resource-limited devices by introducing ZOO, which requires only forward processing and bypasses the resource-intensive BP optimization. 
Concurrently, to mitigate the high-dimensional optimization difficulty caused by the side effect of ZOO, we incorporate model merging of the shared multiple models and set the merge coefficients as the optimization objective, which successfully reduces the optimization dimension. 
In addition, to enhance the synergistic combination of ZOO and model merging, we propose a unique preprocessing strategy that trims intra-model non-influential weights and reduces the inter-model information redundancy. We empirically confirmed the effectiveness of our method using common corruption and style-transferred image benchmarks.
\end{abstract}
\section{Introduction}

\label{sec:intro}
Deep neural networks (DNNs) are powerful tools that have brought remarkable achievements in various fields. However, they often suffer from the domain shift problem, where DNNs degrade their accuracy significantly when training and test data have a large domain gap. 
This issue is nearly impossible to circumvent, given the sheer variety of deployment environments, especially for edge devices in the field.
It is particularly critical in real-world applications, such as automated surveillance systems, quality assurance in manufacturing, and personalized healthcare with wearable devices~\cite{surveillance_DA, automanufacturing_DA, HAR_wearable}. Researchers have long been tackling this problem in the form of domain adaptation and generalization~\cite{domain_adaptation, DGsurvey}. 

\begin{figure*}[t]
    \centering
    \includegraphics[width=\linewidth]{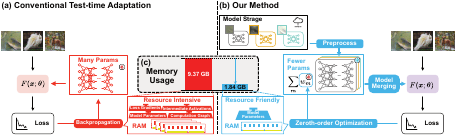}
    \vspace{-5mm}
    \caption{(a) Conventional test-time adaptation (TTA) methods usually update the model parameters through the backpropagation (BP) process. However, this can be disadvantageous because BP requires large memory resources, which limit its applicability to resource-limited scenarios, such as edge computing or low-end smartphones. (b) In this work, we connect model-merging and zeroth-order optimization (ZOO) with a cross-device collaborative TTA framework to simultaneously reduce the memory usage and mitigate the high-dimensional optimization difficulty.}
    \vspace{-5mm}
    \label{fig:teaser}
\end{figure*}

Recently, test-time adaptation (TTA)~\cite{tta_survey} has attracted significant attention.
In contrast to the domain adaptation and generalization approaches that update the model via offline training, TTA updates the model on the fly using only input test data, which are aligned with real-world deployments.
The pioneering approach of TTA is Tent~\cite{Tent}, which updates parts of the model weights so as to minimize the entropy of the model's predictions on test data. Following this, there have been various attempts to extend the applicability of TTA to real-world environments, such as addressing the continual and non-stationary domain shifts~\cite{CoTTA, NOTE, RMT, RoTTA, SAR, ROID}, detecting and removing the noisy samples in the input data stream~\cite{UniEnt, SoTTA}, and developing TTA methods tailored for vision-language models (VLMs)~\cite{TPT, DiffTPT, C-TPT, PromptAlign, SwapPrompt}.

However, most previous TTA methods share a critical limitation: they depend on resource-intensive backpropagation (BP), for which a large amount of memory is particularly indispensable (\cref{fig:teaser} (a)).
Specifically, updating the model via BP requires significantly more memory space than simple forward inference, due to the need to store a computational graph, the activations of the middle layers, the loss gradients of the parameters, etc.
This presents a significant limitation in resource-limited scenarios, i.e., deployment on edge devices like mobile phones and IoT sensors.
These devices frequently encounter domain shift problems due to the uncertain nature of the deployment environments, thus leading to the high demand for solutions with TTA.
Yet, the computational resources of those edge devices are often limited, and thus, existing BP-based algorithms become \textit{infeasible}. The primary research focus of this paper is to develop a resource-friendly TTA method that is \textit{executable} on resource-limited devices, especially with small memory capacity.

This work tackles the aforementioned problem based on the recently established cross-device collaborative TTA (CDC-TTA) framework~\cite{CoLA}. CDC-TTA allows the sharing of the model information among a mixture of resource-abundant and resource-limited devices through a central server, thereby enhancing the overall effectiveness of the service. CDC-TTA is a feasible framework because, in the real-world deployment scenario of TTA, models are usually provided across multiple devices, each of which has different computational resources. On this framework, we build a novel TTA method designed for resource-limited devices (\cref{fig:teaser} (b)), which incorporates the following three pillars.
The first pillar is to switch from the standard BP-based optimization strategy to zeroth-order optimization~\cite{ZO-SGD}. 
Different from BP-based approaches, ZOO can perform optimization only with forward inferences, making it feasible for resource-limited devices that can barely afford inference processing. Unfortunately, ZOO has a drawback in terms of high-dimensional optimization. TTA methods usually optimize a large number of model weights, so simply replacing BP with ZOO results in slow convergence speed and little performance improvement. To mitigate this, we incorporate a model-merging strategy~\cite{SWA} as our second pillar. 
Thanks to the CDC-TTA framework, we can utilize model information from multiple devices, so we build a target model by merging these models. The core insight is to optimize only the merging coefficients, which act as a low-dimensional recipe for how the models are combined. By optimizing this small number of coefficients instead of a vast number of model weights directly, we drastically reduce the dimensionality of the optimization problem, leveraging the benefits of ZOO.
Moreover, the integration of ZOO and model merging, unlike the ad-hoc model construction of existing CDC-TTA methods, can yield a higher accuracy model through training-based model construction.
The final pillar is model weight preprocessing. Directly merging raw model weights may hinder the potential of our method due to uninformative weight values contained in each model and redundant weights overlapping across multiple models. We therefore designed a unique preprocessing to remove these uninformative weights and reduce inter-model redundancy through low-rank approximation.
Our experimental validation of the proposed method on three corruption datasets (CIFAR10-C, CIFAR100-C, and ImageNet-C) and a style transfer dataset (Stylized-ImageNet) confirmed its effectiveness. Notably, the memory usage of the proposed method was confirmed to be significantly lower than that of existing BP-based TTA (\cref{fig:teaser} (c)). Additionally, we investigate the applicability of our method to quantized models, where the BP is inherently inexecutable, and we empirically confirmed that our method can improve the prediction accuracy of the quantized models.

Our main contributions in this work are as follows. 
\begin{itemize}
\item Motivated by the synergy between ZOO and model merging, we propose a new TTA method feasible on resource-limited devices, which optimizes a reasonably small number of merging coefficients via resource-friendly ZOO.
\item We propose a unique preprocessing technique that reduces the interference by suppressing the uninformative model weights in each model and mitigates inter-model weight redundancy by using low-rank approximation, leading to more precise and resource-friendly adaptation.
\end{itemize}
\section{Related Work}
\label{sec:related}

\smallskip
\nbf{Test-time adaptation (TTA)} is an approach that improves a model's inference accuracy by using the input test data. Tent~\cite{Tent}, a pioneering TTA method, achieved annotation-free adaptation to test data by updating the affine parameters of normalization layers~\cite{batch_norm, layer_norm} within the model, utilizing the entropy of the test data predictions as a loss function. Many subsequent TTA methods have been proposed as improvements upon Tent, one direction of which is the extension to more realistic problem settings. \cite{CoTTA} focuses on adapting to continually changing test distributions. ~\cite{NOTE, RMT, RoTTA, SAR, ROID} further extend this problem to handle temporally correlated or dynamic distribution shifts. \cite{UniEnt, SoTTA} are designed to detect out-of-distribution data and exclude them from the adaptation process, and \cite{EATA, EcoTTA} aims to reduce the computational cost for adaptation.
With the recent advancement of Vision-Language Models (VLMs), numerous TTA methods have also been proposed specifically for VLMs~\cite{TPT, DiffTPT, C-TPT, PromptAlign, SwapPrompt, DynaPrompt, TAPT, TTLoRA}. While these techniques significantly improve a model's recognition performance in the test domain, they require resource-intensive BP to update model weights. This poses a significant drawback, as it makes them infeasible for resource-limited environments.

A few TTA methods can perform without BP.
Early approaches often focused on modifying the statistics of batch normalization layers~\cite{BN, RevisitBN, DUA, DELTA, MemBN}, and there are prototype-based methods~\cite{T3A, TAST, AdaNPC} that update class prototypes in the feature space, rather than the model weights themselves. More recent studies~\cite{FOA, FOZO} have introduced BP-free optimization algorithms into TTA for Vision Transformer with prompt tuning~\cite{VPT}. Among TTA for VLMs, several BP-free approaches have been proposed, including feature cache methods~\cite{TDA, ETTA}, an iterative score optimization based on multiple augmented images~\cite{MTA}, and others~\cite{free-on-the-fly, just-shift-it}. Diffusion models~\cite{DDA, SDA} reduce the distribution gap by target-to-source image translation.
While these methods seem suitable for resource-limited devices, we found they have several limitations, such as limited performance improvement due to the rigidity of the frozen model weights, restrictions on applicable model architectures, or a computationally expensive denoising process. 

In this work, we build a TTA method that is executable on resource-limited devices and agnostic to the network architectures based on the cross-device collaborative TTA (CDC-TTA) framework~\cite{CoLA}. The most relevant prior work is CoLA~\cite{CoLA}, which combines multiple models based on the feature distribution similarity. In contrast to such an ad-hoc merging strategy, our method is a training-based approach and yields more precise predictions.

\smallskip
\nbf{Zeroth-order Optimization (ZOO)}, also known as derivative-free optimization, is an approach that optimizes an objective function without using derivative information~\cite{DFO_ZOO}. Its advantage is that it can be applied even when the derivative of the objective function is very difficult or impossible to calculate, or when the function is a black box. 
These derivative-free approaches also meet the demand of the TTA scenarios that need to avoid resource-intensive BP computations due to device limitations.
Several types of ZOO algorithms have been proposed, including direct search (e.g., pattern search~\cite{direct_search}, Nelder-Mead method~\cite{nelder_mead}), model-based search (e.g., Bayesian optimization~\cite{baysian_optimization}), and stochastic/heuristic methods (e.g., Evolution Strategies (ES)~\cite{evolutionary_strategy}, Particle Swarm Optimization (PSO)~\cite{particle_swarm_optimization}, Zeroth-order SGD (ZO-SGD)~\cite{ZO-SGD}). However, ZOO has the weaknesses of slow convergence speed under high-dimensional optimization problems, posing challenges for its direct application to TTA. To address this, we focus on the synergic combination of ZOO and model merging to reduce the dimensionality and develop a TTA method effective even under short adaptation periods.

\smallskip
\nbf{Model Merging} constructs a strong model by combining the weights of multiple pre-trained models. The most popular approach is to simply average the parameters of the models~\cite{SWA}. Several subsequent methods have improved this approach by introducing constraints during the merging process~\cite{FisherMerging, RegMean}. Task arithmetic~\cite{TA} introduced the concept of the ``task vector'', which is the difference between the weights of a base model and the weights of the same model after fine-tuning.
It demonstrated that acquiring and forgetting tasks can be achieved through the addition and subtraction of these task vectors. Other methods have subsequently been proposed, such as mitigating parameter interference between tasks~\cite{TIES, DARE, Model-Breadcrumbs} or applying task arithmetic to LoRA models~\cite{KNOTS}. 
A data-driven method for obtaining merging coefficients has also been proposed~\cite{AdaMerging}. In this work, we conceptualize the framework of cross-device collaborative TTA~\cite{CoLA}, where models are shared among multiple devices, from the perspective of model merging, and leverage the benefit of model merging to overcome the weaknesses of ZOO in high-dimensional optimization.
\section{Preliminaries}
\label{sec:preliminaries}

\smallskip
\nbf{Problem Setup.}
We follow the problem setup of CDC-TTA~\cite{CoLA}. It is a framework that enables a heterogeneous group of devices, comprising both resource-limited and resource-abundant devices, to collaboratively achieve TTA by sharing model weights. While resource-abundant devices can leverage BP to execute an off-the-shelf TTA method and obtain adapted models, the remaining challenge is how to perform TTA on resource-limited devices with the assistance of models from other devices. We focus on addressing the \textit{latter} challenge.

The formal setup is described below. We are given a base pretrained model $F_{\boldsymbol{\theta}_0}$, several fine-tuned models on different devices $\{F_{\boldsymbol{\theta}_m}\}^M_{m=1}$, and unlabeled test data $\boldsymbol{x}_{T_{k'}} \sim P_T(\boldsymbol{x})$. $F_{\boldsymbol{\theta}_0}$ is trained on the source data set $\mathcal{D}_0 = \{\boldsymbol{x}_{0_k}, y_{0_k}\}^{K_0}_{k=1}$, where $K_0$ is the source data set size, $\boldsymbol{x}_{0_k} \sim P_0(\boldsymbol{x})$ and $P_T(\boldsymbol{x}) \neq P_0(\boldsymbol{x})$. Each fine-tuned model $F_{\boldsymbol{\theta}_m}$ is obtained by using an off-the-shelf TTA algorithm and unlabeled data $\mathcal{D}_m = \{\boldsymbol{x}_{m_k}\}^{K_m}_{k=1}$, where $K_m$ is the number of unlabeled data, $\boldsymbol{x}_{m_k} \sim P_m(\boldsymbol{x})$. Note that we are given data only from the test distribution $P_T(\boldsymbol{x})$, but we cannot access the other data, which reflects some real-world limitations (e.g., data privacy). The goal is to minimize the prediction error for the test data. Since we have only unlabeled test data, we usually design an unsupervised loss $\mathcal{L}$ (e.g., prediction entropy) for the model updates. To be more specific, we update the model based on the objective $\min_{\boldsymbol{\theta}_l} \mathcal{L}(\boldsymbol{\theta}_f, \boldsymbol{\theta}_l, \mathcal{B})$ with given test data mini-batch $\mathcal{B}=\{\boldsymbol{x}_{T_{k}}\}^{B}_{k=1}$, where $B$ denotes a batch size, $\boldsymbol{\theta}_f$ denotes the frozen weights and $\boldsymbol{\theta}_l$ denotes the learnable weights, i.e., $\boldsymbol{\theta}=[\boldsymbol{\theta_f};\boldsymbol{\theta_l}]$. In this work, we focus on the TTA problem for resource-limited devices, and thus, we assume that we cannot use resource-intensive BP for model updates.

\smallskip
\nbf{Zeroth-order Optimization (ZOO).} Our method employs zeroth-order SGD (ZO-SGD)~\cite{ZO-SGD}, one of the ZOO algorithms, due to its simplicity and ease of implementation. In ZO-SGD, we obtain pseudo-gradients by Simultaneous Perturbation Stochastic Approximation (SPSA)~\cite{SPSA}. Specifically, SPSA first calculates the loss difference between the two forward passes with positive and negative perturbations as
\begin{equation}
    S = \frac{\mathcal{L}(\boldsymbol{\phi}^+; \mathcal{B}) - \mathcal{L}(\boldsymbol{\phi}^-; \mathcal{B})}{2\delta},
\end{equation}
where $\delta$ is the perturbation size, $\mathcal{B}$ is a mini-batch, and $\boldsymbol{\phi}^+$ and $\boldsymbol{\phi}^-$ are obtained by applying positive and negative perturbations to the original parameter $\boldsymbol{\phi}$ using a random perturbation vector $\boldsymbol{z} \sim \mathcal{N}(0, I_d)$. Then, we update the parameters as
\begin{equation}
    \boldsymbol{\phi} \leftarrow \boldsymbol{\phi} - \eta S\boldsymbol{z},
\end{equation}
where $\eta$ is the learning rate parameter. 

\smallskip
\nbf{Model Merging.} 
We basically follow one of the most popular model merging approaches, task arithmetic~\cite{TA}.
It defines the weight difference between the base model and the fine-tuned model as a ``task vector'' represented as $\boldsymbol{\tau}_m = \boldsymbol{\theta}_m - \boldsymbol{\theta}_0$ where $\boldsymbol{\theta}_m$ and $\boldsymbol{\theta}_0$ are the model weights of the fine-tuned model and the base model, respectively. Utilizing the task vectors from the multiple trained models, we obtain the merged model weights as 
\begin{equation}
    \boldsymbol{\theta}_\text{Merge} = \boldsymbol{\theta}_0 + \sum_m w_m \boldsymbol{\tau}_m, \label{eq:model-wise-merge}
\end{equation}
where $w_m$ denotes the merging coefficients. 

Since different model weights within a DNN (e.g., different layers) learn different information, assigning the same coefficient to all weights of the model could reduce the efficacy of the model merging. Referring to AdaMerge~\cite{AdaMerging}, we prepare a different merging coefficient for each model weight group. Specifically, when the $m$-th model has $N$ model weight groups as $\{\boldsymbol{\theta}^{(n)}_m\}^N_{n=1}$, we assign merging coefficients $w^{(n)}_m$ for the $n$-th model weight group in the $m$-th model, and the merging procedure is represented as
\begin{equation}
    \boldsymbol{\theta}^{(n)}_\text{Merge} = \boldsymbol{\theta}^{(n)}_0 + \sum_m w^{(n)}_m \boldsymbol{\tau}^{(n)}_m, \label{eq:weight_wise_merge}
\end{equation}
where $\boldsymbol{\theta}^{(n)}_0$ is the $n$-th model weight group of the base model and $\boldsymbol{\tau}^{(n)}_m$ is the task vector calculated as $\boldsymbol{\tau}^{(n)}_m = \boldsymbol{\theta}^{(n)}_m - \boldsymbol{\theta}^{(n)}_0$.
\section{Method}
\label{sec:method}

\begin{figure*}[t]
    \begin{center}
    \includegraphics[width=\linewidth]{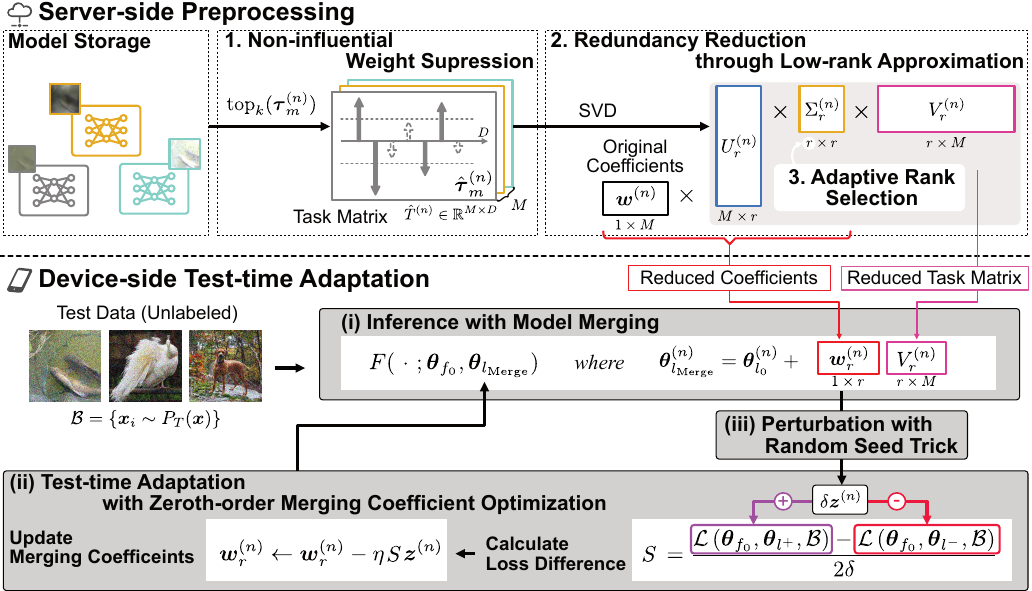}
    \end{center}
    \vspace{-5mm}
    \caption{\textbf{Overview of Our Method.} On the server side, before providing the model weights to each device, preprocessing procedures are applied to reduce the non-influential values and redundancy in the model weights. On the device side, we build the merged model based on the provided initial merging coefficients and task matrices, and we update the coefficients via ZO-SGD.}
    \vspace{-5mm}
    \label{fig:method_overview}
\end{figure*}

\cref{fig:method_overview} presents the overview of our method.
It is built upon the CDC-TTA framework, where the trained models are shared across multiple devices. We construct the target model via model merging using the shared model and achieve TTA on the target device by optimizing the merging coefficients with ZOO. Our method is mainly composed of 1) server-side preprocessing before providing the model weights, and 2) device-side test-time adaptation to adapt the model to the target domain.

\subsection{Server-side Preprocessing}\label{subsec:server-size}

Before providing the model weights to each resource-limited device, we apply the unique preprocessing technique with the following three sub-components to trim the intra-model non-influential weights and to reduce the inter-model information redundancy.

\smallskip
\nbf{1. Non-influential Weight Suppression (NWS).}
We first preprocess the task vectors $\boldsymbol{\tau}^{(n)}_m$ to retain only the task-influential values. As reported in \cite{TIES, Model-Breadcrumbs}, each task vector contains values that do not have effects on merging and may in fact interfere with other task vectors. To mitigate this issue, we apply top-$k$\% filtering for each task vector, setting less influential elements to zero.
Formally, the top-$k$\% filtering is represented as
\begin{equation}
    \hat{\tau}^{(n)}_{m_i} = \left\{
    \begin{array}{ll}
         \tau^{(n)}_{m_i} & (|\tau^{(n)}_{m_i}| \geq \tau^{(n)}_{m_\text{top-$k$\%}}), \\
         0 & (\text{otherwise}).
    \end{array}
    \right.
\end{equation}
We denote this process as $\hat{\boldsymbol{\tau}}^{(n)}_m = \text{top}_k(\boldsymbol{\tau}^{(n)}_m)$.

\smallskip
\nbf{2. Redundancy Reduction (RR) through Low-rank Approximation.}
Next, we focus on the redundancy among the processed task vectors. Intuitively, task vectors that come from similar domains will also be similar, so assigning different coefficients to these task vectors is redundant; it is more efficient to assign a coefficient to the common element between these vectors. Based on this observation, we reduce such redundancy by applying a low-rank approximation. 

The detailed description of the redundancy reduction process is as follows. We re-formulate Eq.~\eqref{eq:weight_wise_merge}\footnote{The subscript $_l$ is used to explicitly indicate that this process needs only to be applied to the learnable weights $\boldsymbol{\theta}_l$.} as
\begin{equation}
    \boldsymbol{\theta}^{(n)}_{l_\text{Merge}} = \boldsymbol{\theta}^{(n)}_{l_0} + \boldsymbol{w}^{(n)}\hat{T}^{(n)},
\end{equation}
where $\boldsymbol{w}^{(n)} \in \mathbb{R}^M$ is the coefficient vector, $\hat{T}^{(n)} \in \mathbb{R}^{M \times D}$ is the task matrix obtained by concatenating the task vectors $\{\hat{\boldsymbol{\tau}}^{(n)}_m\}^{M}_{m=1}$, and $D$ is the size of each task vector. Then, we apply singular value decomposition (SVD) to the task matrix $\hat{T}^{(n)}$ and obtain a low-rank approximated version
\begin{equation}
    \hat{T}^{(n)}_r = U^{(n)}_r \Sigma^{(n)}_r V^{(n)}_r,
\end{equation}
where $U^{(n)}_r \in \mathbb{R}^{M \times r}$, $\Sigma^{(n)}_r \in \text{Diag}(\mathbb{R}^r)$, and $V^{(n)}_r \in \mathbb{R}^{r \times D}$. Finally, our model-merging procedure is represented as
\begin{align}
    \boldsymbol{\theta}^{(n)}_{l_\text{Merge}} & = \boldsymbol{\theta}^{(n)}_{l_0} + \boldsymbol{w}^{(n)} U^{(n)}_r \Sigma^{(n)}_r V^{(n)}_r, \\
    & = \boldsymbol{\theta}^{(n)}_{l_0} + \boldsymbol{w}^{(n)}_r V^{(n)}_r,
\end{align}
where $\boldsymbol{w}^{(n)}_r = \boldsymbol{w}^{(n)} U^{(n)}_r \Sigma^{(n)}_r \in \mathbb{R}^r$. 

\smallskip
\nbf{3. Adaptive Rank Selection (ARS).}
As mentioned previously, each model weight in a neural network learns different information. Consequently, the similarity trend of the task vector varies for each model weight, and the appropriate rank for the above low-rank approximation procedure also differs for each. 
To be more specific, if the weights are similar across models, an unnecessarily high rank will undermine the effectiveness of redundancy reduction. Conversely, if the weights differ significantly between models, setting the rank too low will result in a loss of information in the task matrix.
To address this, we introduce a technique to automatically select the rank for each model weight, further reducing the total number of merging coefficients to be optimized. In this technique, we focus on the recovery error of the original task matrix. The recovery error between the original task matrix and the low-rank approximated one is defined as
\begin{equation}
    R(\hat{T}^{(n)}, r) = \frac{\left\|\hat{T}^{(n)}-\hat{T}^{(n)}_r\right\|_F^2}{\|\hat{T}^{(n)}\|_F^2}.
\end{equation}
By adjusting the rank so that the recovery error is small enough, we can select the appropriate rank for each model weight. Specifically, the minimum necessary rank is set based on the following equation:
\begin{equation}
    r^{(n)} = \min_r \left\{r | R(\hat{T}^{(n)}, r) < \epsilon \times D \right\}, \label{eq:adaptive_rank_selection}
\end{equation}
where $\epsilon$ is a parameter to tune the recovery error. In addition, we limit the maximum size of the rank as $r^{(n)} \leftarrow \min(r^{(n)}, r_\text{max})$.

The combination of RR and ARS offers another important advantage; we can reduce the dimensionality of the optimization. As discussed earlier, ZOO methods usually struggle with the difficulty of a high-dimensional optimization. Our method mitigates this by introducing a model-merging strategy instead of directly optimizing the model weights. However, as the number of merging models increases, the number of merging coefficients also increases linearly, which re-ignites the aforementioned issue. Our approach contributes to reducing the optimization dimensionality by grouping similar task vectors together, thereby virtually decreasing the number of merging models. Specifically, our method can reduce the number of merging coefficients by $r/M$.

The overall preprocessing procedure is outlined in Algorithm~\ref{alg:preprocess} in the supplementary material.
Note that this is entirely conducted on the high-power server-side systems, so there is no need for concerns regarding the high computational costs of SVD, and it can be reduced by employing randomized SVD~\cite{randomizedSVD}.

\subsection{Device-side Test-time Adaptation}\label{subsec:device-size}

Even when training a subset of the model weights, BP requires significant memory to store activations of the intermediate layers and the whole computational graph, making it difficult to perform on resource-limited devices. On the other hand, the direct application of ZO-SGD causes poor convergence under high-dimensional optimization problems. Nevertheless, our method successfully reduces the dimensional size by introducing model merging and the preprocessing in \cref{subsec:server-size}, taking full advantage of ZO-SGD. The concrete procedure of our device-side test-time adaptation is as follows.

\smallskip
\nbf{(i) Inference with Model Merging.}
Given reduced merging coefficients $\{\boldsymbol{w}^{(n)}_r\}^{N}_{n=1}$  and reduced task matrices $\{V^{(n)}_r\}^{N}_{n=1}$, we calculate the merged model weights as
\begin{equation}
    \boldsymbol{\theta}^{(n)}_{l_\text{Merge}} = \boldsymbol{\theta}^{(n)}_{l_0} + \boldsymbol{w}^{(n)}_r V^{(n)}_r. \label{eq:tta_merge}
\end{equation}

\smallskip
\nbf{(ii) Zeroth-order Merging Coefficient Optimization.}
We first calculate the loss difference between the positively and negatively perturbed merging coefficients. When we obtain perturbations $\boldsymbol{z}^{(n)} \sim \mathcal{N}(0, I_{r^{(n)}})$, we add and subtract them from the merging coefficients $w^{(n)}_r$ and get perturbed model weights $\boldsymbol{\theta}^{(n)}_{l^+}$ and $\boldsymbol{\theta}^{(n)}_{l^-}$ based on Eq.~\eqref{eq:tta_merge}, respectively. Then, we calculate the loss difference as 
\begin{equation}
    S = \text{clamp}_{[-c_\text{clamp}, c_\text{clamp}]}\left(\frac{\mathcal{L}\left(\boldsymbol{\theta}_{f_0}, \boldsymbol{\theta}_{l^+}, \mathcal{B}\right) - \mathcal{L}\left(\boldsymbol{\theta}_{f_0}, \boldsymbol{\theta}_{l^-}, \mathcal{B}\right)}{2\delta}\right),
\end{equation}
where we apply the clamp function to the original difference in order to reduce the impact of overly large loss differences. Finally, we calculate a pseudo-gradient and utilize it to update the merging coefficients, as
\begin{equation}
    \boldsymbol{w}^{(n)}_r \leftarrow \boldsymbol{w}^{(n)}_r - \eta S \boldsymbol{z}^{(n)}.
\end{equation}

\smallskip
\nbf{(iii) Random Seed Trick for Memory Usage Suppression.}
We reduce the memory usage of the merge coefficient optimization process by using the random seed trick~\cite{MeZO}. Specifically, instead of storing the random perturbations $\{\boldsymbol{z}^{(n)}\}^N_{n=1}$, we sample a random seed $s$ for each iteration and resample the random perturbations. This trick makes the memory cost required for the optimization process almost equivalent to the normal inference process, which makes our method suitable for TTA on resource-limited devices. The detailed procedure is described in Algorithm~\ref{alg:zo_sgd} in the supplementary material.
\section{Experiments}
\label{sec:experiments}

We experimentally validated our method on common corruption benchmarks and style transfer benchmarks and conducted comprehensive analyses of our method.

\subsection{Setups}

\smallskip
\nbf{Datasets.} We utilized three datasets for common corruption benchmarks: CIFAR10-C, CIFAR100-C, and ImageNet-C~\cite{Corruptions}.
They were generated by applying corruptions to the original CIFAR-10, CIFAR-100~\cite{CIFAR}, and ImageNet-1k~\cite{Imagenet} datasets, respectively. Specifically, 15 types of corruption categorized into four groups (noise, blur, weather, and digital) were applied to the validation split of each original dataset. In the TTA evaluation, a model pre-trained on the original training split is updated and tested on the corrupted validation data.

We conducted a benchmark evaluation on style-transferred images to assess performance under other types of domain shifts. The style-transferred images were generated following Stylized-ImageNet~\cite{StylizedImagenet}. Specifically, the ImageNet validation data was transformed by a style-transferring method~\cite{AdaIN} according to various style-providing images. For our experiments, we prepared a total of 50 different style-transferred datasets. To evaluate the performance across a varying number of domains, we set up five experimental conditions with 10, 20, 30, 40, and 50 domains, conducting evaluations for each.

\begin{table*}[t]
\centering
\caption{\textbf{Experimental results on CIFAR10-C and CIFAR100-C.} The evaluation results are presented in terms of error rate [\%]. The best are highlighted in \textbf{bold}.
}
\footnotesize
\vspace{-3mm}
\begin{tabular}{wl{15mm}wc{3.5mm}wc{3.5mm}wc{3.5mm}wc{3.5mm}wc{3.5mm}wc{3.5mm}wc{3.5mm}wc{3.5mm}wc{3.5mm}wc{3.5mm}wc{3.5mm}wc{3.5mm}wc{3.5mm}wc{3.5mm}wc{3.5mm}wc{3.5mm}}
\toprule[0.3mm]
\multicolumn{17}{c}{CIFAR10-C} \\
\midrule
    & \multicolumn{3}{c}{Noise}
    & \multicolumn{4}{c}{Blur}
    & \multicolumn{3}{c}{Weather}
    & \multicolumn{5}{c}{Digital}
    & Avg.\\
      
      Method 
    & Gaus. 
    & Shot 
    & Impu. 
    & Defo. 
    & Glas. 
    & Moti. 
    & Zoom 
    & Snow
    & Fros.
    & Fog
    & Brig.
    & Cont.
    & Elas.
    & Pixe.
    & Jpeg
    & \\
\cmidrule(r{0.5mm}){1-1}\cmidrule(l{0.5mm}r{0.5mm}){2-4}\cmidrule(l{0.5mm}r{0.5mm}){5-8}\cmidrule(l{0.5mm}r{0.5mm}){9-11}\cmidrule(l{0.5mm}r{0.5mm}){12-16}\cmidrule(l{0.5mm}){17-17}
      No Adapt 
    & 72.3
    & 65.7
    & 72.9
    & 46.9
    & 54.3
    & 34.8
    & 42.0
    & 25.1
    & 41.3
    & 26.0
    & 9.3
    & 46.7
    & 26.6
    & 58.5
    & 30.3
    & 43.5 \\
      BN stats
    & 28.3
    & 26.0
    & 36.2
    & 12.6
    & 34.9
    & 13.9
    & 12.0
    & 17.5
    & 17.6
    & 14.9
    & 8.2
    & 13.0
    & 23.5
    & 19.5
    & 27.2
    & 20.3 \\
      T3A
    & 61.7
    & 56.5
    & 67.4
    & 41.0
    & 47.5
    & 31.7
    & 36.2
    & 24.2
    & 37.1
    & 24.2
    & 9.4
    & 42.4
    & 25.3
    & 49.0
    & 29.5
    & 38.9 \\
\cmidrule(r{0.5mm}){1-1}\cmidrule(l{0.5mm}r{0.5mm}){2-4}\cmidrule(l{0.5mm}r{0.5mm}){5-8}\cmidrule(l{0.5mm}r{0.5mm}){9-11}\cmidrule(l{0.5mm}r{0.5mm}){12-16}\cmidrule(l{0.5mm}){17-17}
      CoLA (DeYO)
    & 25.3
    & 22.9
    & 33.2
    & 11.8
    & 32.4
    & 13.0
    & 11.5
    & 15.9
    & 15.6
    & 13.7
    & 7.8
    & 11.1
    & 21.8
    & 17.4
    & 24.4
    & 18.5 \\
      \rowcolor[gray]{0.9}\textbf{Ours} (DeYO)
    & 24.8
    & 22.2
    & 32.8
    & 11.8
    & 32.0
    & 13.0
    & 11.5
    & 15.6
    & 15.3
    & 13.7
    & 7.8
    & \textbf{10.9}
    & 21.7
    & 17.1
    & 23.9
    & 18.3 \\
\cmidrule(r{0.5mm}){1-1}\cmidrule(l{0.5mm}r{0.5mm}){2-4}\cmidrule(l{0.5mm}r{0.5mm}){5-8}\cmidrule(l{0.5mm}r{0.5mm}){9-11}\cmidrule(l{0.5mm}r{0.5mm}){12-16}\cmidrule(l{0.5mm}){17-17}
      CoLA (ROID)
    & 22.9
    & 20.5
    & 29.8
    & \textbf{11.1}
    & 29.4
    & \textbf{12.3}
    & \textbf{10.5}
    & 14.9
    & 14.8
    & 13.2
    & \textbf{7.6}
    & 11.3
    & 19.8
    & 15.4
    & 21.7
    & 17.0 \\
      \rowcolor[gray]{0.9}\textbf{Ours} (ROID)
    & \textbf{20.7}
    & \textbf{18.7}
    & \textbf{27.4}
    & 11.2
    & \textbf{27.9}
    & 12.7
    & 10.6
    & \textbf{14.8}
    & \textbf{14.6}
    & \textbf{12.8}
    & 7.9
    & 11.3
    & \textbf{19.3}
    & \textbf{14.4}
    & \textbf{19.7}
    & \textbf{16.3} \\
\midrule[0.3mm]
    \multicolumn{17}{c}{CIFAR100-C} \\
\midrule
    & \multicolumn{3}{c}{Noise}
    & \multicolumn{4}{c}{Blur}
    & \multicolumn{3}{c}{Weather}
    & \multicolumn{5}{c}{Digital}
    & Avg.\\
      
      Method 
    & Gaus. 
    & Shot 
    & Impu. 
    & Defo. 
    & Glas. 
    & Moti. 
    & Zoom 
    & Snow
    & Fros.
    & Fog
    & Brig.
    & Cont.
    & Elas.
    & Pixe.
    & Jpeg
    & \\
\cmidrule(r{0.5mm}){1-1}\cmidrule(l{0.5mm}r{0.5mm}){2-4}\cmidrule(l{0.5mm}r{0.5mm}){5-8}\cmidrule(l{0.5mm}r{0.5mm}){9-11}\cmidrule(l{0.5mm}r{0.5mm}){12-16}\cmidrule(l{0.5mm}){17-17}
      No Adapt 
    & 73.0
    & 68.0
    & 39.4
    & 29.4
    & 54.1
    & 30.8
    & 28.8
    & 39.5
    & 45.8
    & 50.3
    & 29.5
    & 55.1
    & 37.2
    & 74.7
    & 41.2
    & 46.4 \\
      BN stats
    & 42.3
    & 40.8
    & 43.3
    & 27.8
    & 42.0
    & 29.8
    & 27.9
    & 35.1
    & 35.0
    & 41.7
    & 26.3
    & 30.3
    & 35.6
    & 33.4
    & 41.3
    & 35.5 \\
      T3A
    & 72.1
    & 66.8
    & 39.3
    & 29.1
    & 50.7
    & 30.8
    & 28.2
    & 39.8
    & 45.0
    & 49.2
    & 29.6
    & 51.5
    & 36.3
    & 70.7
    & 41.2
    & 45.3 \\
\cmidrule(r{0.5mm}){1-1}\cmidrule(l{0.5mm}r{0.5mm}){2-4}\cmidrule(l{0.5mm}r{0.5mm}){5-8}\cmidrule(l{0.5mm}r{0.5mm}){9-11}\cmidrule(l{0.5mm}r{0.5mm}){12-16}\cmidrule(l{0.5mm}){17-17}
      CoLA (DeYO)
    & 35.4
    & 33.8
    & 37.5
    & 24.7
    & \textbf{35.6}
    & \textbf{26.5}
    & \textbf{24.3}
    & \textbf{30.6}
    & \textbf{29.7}
    & 35.3
    & 23.8
    & \textbf{26.7}
    & \textbf{31.0}
    & 28.0
    & 36.8
    & 30.7 \\
      \rowcolor[gray]{0.9}\textbf{Ours} (DeYO)
    & 35.4
    & 33.9
    & 37.2
    & 25.0 
    & \textbf{35.6}
    & 26.9
    & 24.9
    & 31.2
    & 30.3
    & \textbf{35.1}
    & 24.2
    & 27.2
    & 31.2
    & 28.1
    & 37.1
    & 30.9 \\
\cmidrule(r{0.5mm}){1-1}\cmidrule(l{0.5mm}r{0.5mm}){2-4}\cmidrule(l{0.5mm}r{0.5mm}){5-8}\cmidrule(l{0.5mm}r{0.5mm}){9-11}\cmidrule(l{0.5mm}r{0.5mm}){12-16}\cmidrule(l{0.5mm}){17-17}
      CoLA (ROID)
    & 37.1
    & 35.4
    & 38.4
    & 25.0
    & 37.1
    & 27.1
    & 24.9
    & 31.1
    & 30.8
    & 37.3
    & 24.0
    & 27.5
    & 31.8
    & 29.0
    & 37.5
    & 31.6 \\
      \rowcolor[gray]{0.9}\textbf{Ours} (ROID)
    & \textbf{35.2}
    & \textbf{34.1}
    & 35.7
    & \textbf{24.6}
    & 35.7
    & \textbf{26.5}
    & 24.6
    & \textbf{30.6}
    & 30.1
    & 35.3
    & \textbf{23.7}
    & 27.1
    & 31.1
    & \textbf{27.4}
    & \textbf{35.7} 
    & \textbf{30.5} \\
\bottomrule[0.3mm]
\end{tabular}
\vspace{-3mm}
\label{tab:common_corruption_cifar}
\end{table*}

\begin{table*}[t]
\centering
\caption{\textbf{Experimental results on ImageNet-C.} The evaluation results are presented in terms of error rate [\%]. The best are highlighted in \textbf{bold}.
}
\vspace{-3mm}
\footnotesize
\begin{tabular}{wl{15mm}wc{3.5mm}wc{3.5mm}wc{3.5mm}wc{3.5mm}wc{3.5mm}wc{3.5mm}wc{3.5mm}wc{3.5mm}wc{3.5mm}wc{3.5mm}wc{3.5mm}wc{3.5mm}wc{3.5mm}wc{3.5mm}wc{3.5mm}wc{3.5mm}}
\toprule[0.3mm]
\multicolumn{17}{c}{ImageNet-C (ResNet-50)} \\
\midrule
    & \multicolumn{3}{c}{Noise}
    & \multicolumn{4}{c}{Blur}
    & \multicolumn{3}{c}{Weather}
    & \multicolumn{5}{c}{Digital}
    & Avg.\\
      
      Method 
    & Gaus. 
    & Shot 
    & Impu. 
    & Defo. 
    & Glas. 
    & Moti. 
    & Zoom 
    & Snow
    & Fros.
    & Fog
    & Brig.
    & Cont.
    & Elas.
    & Pixe.
    & Jpeg
    & \\
\cmidrule(r{0.5mm}){1-1}\cmidrule(l{0.5mm}r{0.5mm}){2-4}\cmidrule(l{0.5mm}r{0.5mm}){5-8}\cmidrule(l{0.5mm}r{0.5mm}){9-11}\cmidrule(l{0.5mm}r{0.5mm}){12-16}\cmidrule(l{0.5mm}){17-17}
      No Adapt 
    & 97.8 
    & 97.1 
    & 98.2 
    & 81.7 
    & 89.8
    & 85.2
    & 78.0
    & 83.5 
    & 77.1 
    & 75.9 
    & 41.3 
    & 94.6 
    & 82.5 
    & 79.3 
    & 68.6 
    & 82.0 \\
      BN stats
    & 85.0 
    & 84.2 
    & 85.0
    & 85.2
    & 84.4
    & 73.2
    & 61.2 
    & 65.9
    & 68.1
    & 51.9
    & 34.8
    & 83.1
    & 55.9
    & 51.3
    & 59.9
    & 68.6 \\
      T3A
    & 98.4
    & 98.0
    & 98.9
    & 83.7
    & 91.7
    & 87.4
    & 80.4
    & 86.0 
    & 79.6
    & 79.4
    & 41.9
    & 95.7
    & 86.0
    & 82.4
    & 71.9
    & 84.1 \\
\cmidrule(r{0.5mm}){1-1}\cmidrule(l{0.5mm}r{0.5mm}){2-4}\cmidrule(l{0.5mm}r{0.5mm}){5-8}\cmidrule(l{0.5mm}r{0.5mm}){9-11}\cmidrule(l{0.5mm}r{0.5mm}){12-16}\cmidrule(l{0.5mm}){17-17}
      CoLA (DeYO)
    & 74.3
    & 73.5
    & 73.6
    & 76.8
    & 74.8
    & 65.5
    & 56.0
    & 60.5
    & 62.4
    & 49.3
    & 35.6
    & 72.3
    & 49.9
    & 46.0
    & 53.0
    & 61.6 \\
      \rowcolor[gray]{0.9}\textbf{Ours} (DeYO)
    & 72.4
    & 71.3
    & 71.9
    & 76.6
    & 74.0
    & 66.1
    & 56.6
    & 61.0
    & 62.5
    & 50.7
    & 37.2
    & 72.1
    & 49.9
    & 46.3
    & 52.6
    & 61.4 \\
\cmidrule(r{0.5mm}){1-1}\cmidrule(l{0.5mm}r{0.5mm}){2-4}\cmidrule(l{0.5mm}r{0.5mm}){5-8}\cmidrule(l{0.5mm}r{0.5mm}){9-11}\cmidrule(l{0.5mm}r{0.5mm}){12-16}\cmidrule(l{0.5mm}){17-17}
      CoLA (ROID)
    & 74.8
    & 74.5
    & 74.1
    & 77.0
    & 76.0
    & 65.0
    & 55.2
    & 59.3
    & 61.8
    & \textbf{47.4}
    & \textbf{34.2}
    & 72.3
    & 50.2
    & 44.9
    & 52.1
    & 61.3 \\
      \rowcolor[gray]{0.9}\textbf{Ours} (ROID)
    & \textbf{70.4}
    & \textbf{69.7} 
    & \textbf{69.8} 
    & \textbf{74.4} 
    & \textbf{72.8} 
    & \textbf{63.5} 
    & \textbf{54.9} 
    & \textbf{58.9} 
    & \textbf{61.1} 
    & 47.5 
    & 35.8 
    & \textbf{69.2} 
    & \textbf{49.0} 
    & \textbf{44.2} 
    & \textbf{50.2} 
    & \textbf{59.4} \\
\midrule[0.3mm]
\multicolumn{17}{c}{ImageNet-C (ViT B/16)} \\
\midrule
    & \multicolumn{3}{c}{Noise}
    & \multicolumn{4}{c}{Blur}
    & \multicolumn{3}{c}{Weather}
    & \multicolumn{5}{c}{Digital}
    & Avg.\\
      
      Method 
    & Gaus. 
    & Shot 
    & Impu. 
    & Defo. 
    & Glas. 
    & Moti. 
    & Zoom 
    & Snow
    & Fros.
    & Fog
    & Brig.
    & Cont.
    & Elas.
    & Pixe.
    & Jpeg
    & \\
\cmidrule(r{0.5mm}){1-1}\cmidrule(l{0.5mm}r{0.5mm}){2-4}\cmidrule(l{0.5mm}r{0.5mm}){5-8}\cmidrule(l{0.5mm}r{0.5mm}){9-11}\cmidrule(l{0.5mm}r{0.5mm}){12-16}\cmidrule(l{0.5mm}){17-17}
      No Adapt 
    & 65.8 
    & 67.3 
    & 65.3 
    & 68.8 
    & 74.4 
    & 64.3 
    & 66.6 
    & 56.8 
    & 45.2 
    & 48.6 
    & 29.2 
    & 81.8 
    & 57.1 
    & 60.8 
    & 50.2 
    & 60.2 \\
      BN stats
    & 65.8 
    & 67.3 
    & 65.3 
    & 68.8 
    & 74.4 
    & 64.3 
    & 66.6 
    & 56.8 
    & 45.2 
    & 48.6 
    & 29.2 
    & 81.8 
    & 57.1 
    & 60.8 
    & 50.2 
    & 60.2 \\
      T3A
    & 65.8 
    & 67.4
    & 65.3
    & 68.8
    & 74.2
    & 64.4
    & 66.3
    & 56.7
    & 45.4
    & 48.6
    & 29.2
    & 89.5
    & 56.9
    & 60.8
    & 50.1
    & 60.6 \\
\cmidrule(r{0.5mm}){1-1}\cmidrule(l{0.5mm}r{0.5mm}){2-4}\cmidrule(l{0.5mm}r{0.5mm}){5-8}\cmidrule(l{0.5mm}r{0.5mm}){9-11}\cmidrule(l{0.5mm}r{0.5mm}){12-16}\cmidrule(l{0.5mm}){17-17}
      CoLA (DeYO)
    & 62.1
    & 62.8
    & 61.5
    & 64.7
    & 69.9
    & 60.2
    & 62.6
    & 54.6
    & 45.6
    & 45.2
    & 28.0
    & \textbf{73.5}
    & 54.8
    & 56.0
    & 47.0
    & 56.6 \\
      \rowcolor[gray]{0.9}\textbf{Ours} (DeYO)
    & \textbf{60.3}
    & 61.5
    & \textbf{60.3}
    & 64.2
    & 68.3
    & 58.6
    & \textbf{61.2}
    & \textbf{54.4}
    & 47.1
    & 46.1
    & \textbf{27.9}
    & 75.4
    & 53.8
    & 53.3
    & 45.4
    & 55.9 \\
\cmidrule(r{0.5mm}){1-1}\cmidrule(l{0.5mm}r{0.5mm}){2-4}\cmidrule(l{0.5mm}r{0.5mm}){5-8}\cmidrule(l{0.5mm}r{0.5mm}){9-11}\cmidrule(l{0.5mm}r{0.5mm}){12-16}\cmidrule(l{0.5mm}){17-17}
      CoLA (ROID)
    & 63.6
    & 64.8
    & 62.2
    & 64.6
    & 70.7
    & 61.4
    & 64.6
    & 55.3 
    & \textbf{44.8}
    & 50.7
    & \textbf{27.9}
    & 78.4
    & 54.2
    & 56.2
    & 47.1
    & 57.8 \\
      \rowcolor[gray]{0.9}\textbf{Ours} (ROID)
    & 62.0
    & \textbf{61.3}
    & 60.6
    & \textbf{59.5}
    & \textbf{61.8}
    & \textbf{55.0}
    & 62.2
    & \textbf{54.4}
    & 46.6
    & \textbf{44.8}
    & 28.5
    & 74.2
    & \textbf{50.5}
    & \textbf{51.5}
    & \textbf{42.2}
    & \textbf{54.3} \\
\bottomrule[0.3mm]
\end{tabular}
\vspace{-3mm}
\label{tab:common_corruption_imagenet}
\end{table*}

\smallskip
\nbf{Comparison Methods.} 
We compared our method with the following three existing methods that do not use BP: 1) BN stats~\cite{BN}, which reduces the domain shifts by using the mean and variance of the testing mini-batch for the batch-normalization procedure, instead of relying on the stored statistics (i.e., running mean and running variance), 2) T3A~\cite{T3A}, which performs inference based on class prototypes in the feature space and updates these prototypes according to the test data on-the-fly, 3) CoLA~\cite{CoLA}, which is the fairest competitor to ours, as it also builds a target model from multiple models. CoLA combines the models based on the similarities of the image feature statistics.

\smallskip
\nbf{Evaluation Protocol.}
We first prepared the sharing model weights using existing TTA methods. Specifically, for each domain shift, we performed one epoch of adaptation using a TTA method, and the resulting model weights were utilized as the source of model merging. In this work, we used two TTA methods to obtain these source model weights: DeYO~\cite{DeYO} and ROID~\cite{ROID}.

In the evaluation, we measured the classification error rate on each domain shift and assessed the overall performance by the average error rate of all shifts. Note that we \textit{excluded} the model that was adapted to the target domain from the merging model pool. For example, when evaluating performance on the ``Gaussian noise'' data in the common corruption benchmark, we built the merged model from the other 14 models, each adapted to a different corruption.

We use WideResNet-28~\cite{WRN} for CIFAR10-C, ResNeXt-29~\cite{resnext} for CIFAR100-C, ResNet-50~\cite{resnet} for ImageNet-C, and ViT-B/16~\cite{vit} for ImageNet-C and Stylized-ImageNet, following~\cite{ROID}. The other implementation details, including the hyperparameter settings, are described in Sec.~\ref{sec:implementation_details} in the supplementary material.
\subsection{Results}

We evaluated the performance of our method by taking the average score of three different runs for all benchmarks.

\smallskip
\nbf{Common Corruption Benchmarks (\cref{tab:common_corruption_cifar} and \cref{tab:common_corruption_imagenet}).}
The superiority of our method was consistently confirmed across all three datasets. Comparing the methods with and without model merging, we found that the former achieved higher accuracy with large margins. These results support our central hypothesis that model merging is an effective technique for TTA. Comparing our method with CoLA, ours outperformed in most cases. This indicates that it is more effective to optimize the merging coefficient in the training loss-based approach than the non-optimization-based methods, demonstrating the validity of introducing ZOO to TTA. Finally, we investigated the effect of the difference in the TTA methods used for preparing the sharing models (DeYO and ROID). As our algorithm design is agnostic to such selection, the experimental results showed that those differences did not hinder the effectiveness of our method.

\begin{table}[t]
\centering
\caption{
\textbf{Experimental results on Style Transferred benchmarks.} The evaluation results are presented in terms of error [\%]. The best are highlighted in \textbf{bold}.
}
\footnotesize
\vspace{-3mm}
\begin{tabular}{wl{12mm}wc{6mm}wc{6mm}wc{6mm}wc{6mm}wc{6mm}}
\toprule[0.3mm]
    & \multicolumn{5}{c}{Number of Domains} \\
      Method 
    & 10
    & 20
    & 30
    & 40
    & 50 \\
\cmidrule(r{0.5mm}){1-1}\cmidrule(l{0.5mm}){2-6}
      No Adapt
    & 81.1
    & 78.8
    & 80.0
    & 80.4
    & 80.2 \\
      T3A
    & 81.0
    & 78.7
    & 79.8
    & 80.3
    & 80.0 \\
\cmidrule(r{0.5mm}){1-1}\cmidrule(l{0.5mm}){2-6}
      CoLA
    & 77.0
    & 73.1
    & 75.0
    & 75.5
    & 75.2 \\
      \rowcolor[gray]{0.9}\textbf{Ours}
    & \textbf{71.8}
    & \textbf{66.4}
    & \textbf{67.9}
    & \textbf{69.4}
    & \textbf{69.2} \\
\bottomrule[0.3mm]
\end{tabular}
\vspace{-3mm}
\label{tab:style_transfer_results}
\end{table}

\begin{table}[t]
\centering
\caption{
\textbf{Ablation Study.} We evaluate the effectiveness of each component using ImageNet-C and ViT-B/16.}
\footnotesize
\vspace{-3mm}
\begin{tabular}{wl{20mm}wc{6mm}wc{6mm}wc{6mm}wc{18mm}}
\toprule[0.3mm]

    & \multicolumn{3}{c}{Preprocessing}
    & \\
    
    Method
    & NWS
    & RR
    & ARS
    & Error Rate [\%] \\

\cmidrule(r{0.5mm}){1-1}\cmidrule(l{0.5mm}r{0.5mm}){2-4}\cmidrule(l{0.5mm}){5-5}

      No Adapt 
    &
    &
    &
    & 60.2 \\

\cmidrule(r{0.5mm}){1-1}\cmidrule(l{0.5mm}r{0.5mm}){2-4}\cmidrule(l{0.5mm}){5-5}

      \multirow{2}{*}{Model Merging}
    & 
    &
    & 
    & 55.4 \\

    & \checkmark
    &
    & 
    & 55.1 \\

\cmidrule(r{0.5mm}){1-1}\cmidrule(l{0.5mm}r{0.5mm}){2-4}\cmidrule(l{0.5mm}){5-5}

    & \checkmark
    & 
    &
    & 54.8 \\

    Model Merging
    & \checkmark
    & \checkmark
    &
    & 54.4 \\

    + ZO-SGD
    & \checkmark
    &
    & \checkmark
    & 54.7 \\

    & \checkmark
    & \checkmark
    & \checkmark
    & \textbf{54.3} \\

\bottomrule[0.3mm]
\end{tabular}
\vspace{-3mm}
\label{tab:ablation_study}
\end{table}

\smallskip
\nbf{Style-Transferred Benchmarks (\cref{tab:style_transfer_results}).}
Similar to the results on the common corruption benchmark, our method demonstrated its superiority. The performance trend was also consistent with the common corruption benchmark result: specifically, methods utilizing the multiple model weights outperformed those without them. We confirmed that the number of domains used for evaluation did not affect the performance of our method.

\smallskip
\nbf{Discussion.} 
Compared to CoLA, our method incurs higher computational costs. Specifically, while CoLA requires a single forward pass, our approach necessitates two additional forward passes (or more, depending on the loss design) for the update process, resulting in the higher computational overhead. Although this represents a limitation of our method, it does not constitute a critical drawback. Once the adaptation to the target domain has sufficiently progressed, we can recover inference speed by halting the further update process (until another drastic domain shift occurs). While how to implement this may be an important consideration in practical deployments, the optimal solution varies widely depending on application-specific demands. We view this primarily as an engineering challenge rather than the core research problem of this study, i.e, addressing the device resource constraint for optimization processing; therefore, we reserve its detailed investigation for future work.

\subsection{Analysis}

\begin{figure}[t]
    \centering
    \includegraphics[width=1.0\linewidth]{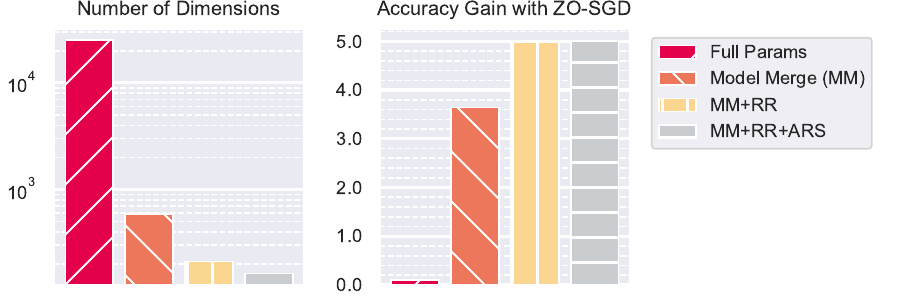}
    \caption{\textbf{Dimension Reduction Analysis.} We examined how much our proposed techniques can reduce dimensionality and the resulting performance gain with ZO-SGD.
    }
    \vspace{-3mm}
    \label{fig:num_param_analysis}
\end{figure}

\begin{figure}[t]
    \centering
    \includegraphics[width=1.0\linewidth]{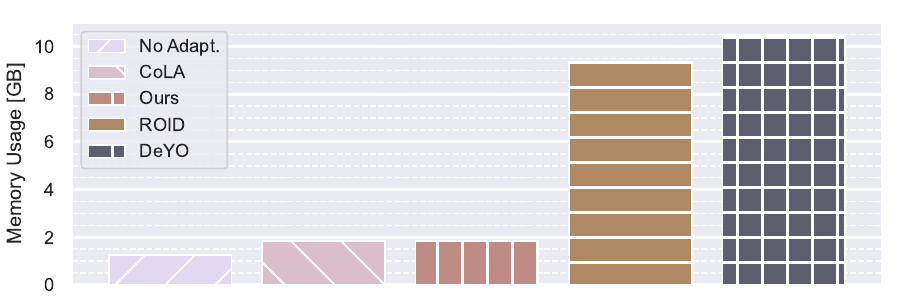}
    \caption{\textbf{Memory Usage Analysis.} We compared the memory usage of CDC-TTA methods (CoLA and Ours), and the conventional BP-based methods (ROID and DeYO).}
    \vspace{-3mm}
    \label{fig:memory_usage}
\end{figure}

\begin{table*}[t]
\caption{\textbf{Analysis of Application to Quantized Model.} The results are presented in error rate [\%]. The best are highlighted in \textbf{bold}.
}
\centering
\footnotesize
\vspace{-3mm}
\begin{tabular}{wl{15mm}wc{2.5mm}wc{2.5mm}wc{2.5mm}wc{2.5mm}wc{2.5mm}wc{2.5mm}wc{2.5mm}wc{2.5mm}wc{2.5mm}wc{2.5mm}wc{2.5mm}wc{2.5mm}wc{2.5mm}wc{2.5mm}wc{2.5mm}wc{2.5mm}}
\toprule[0.3mm]
    & \multicolumn{3}{c}{Noise}
    & \multicolumn{4}{c}{Blur}
    & \multicolumn{3}{c}{Weather}
    & \multicolumn{5}{c}{Digital}
    & Avg.\\
      
      Method 
    & Gaus. 
    & Shot 
    & Impu. 
    & Defo. 
    & Glas. 
    & Moti. 
    & Zoom 
    & Snow
    & Fros.
    & Fog
    & Brig.
    & Cont.
    & Elas.
    & Pixe.
    & Jpeg
    & \\
\cmidrule(r{0.5mm}){1-1}\cmidrule(l{0.5mm}r{0.5mm}){2-4}\cmidrule(l{0.5mm}r{0.5mm}){5-8}\cmidrule(l{0.5mm}r{0.5mm}){9-11}\cmidrule(l{0.5mm}r{0.5mm}){12-16}\cmidrule(l{0.5mm}){17-17}
      No Adapt 
    & 70.6
    & 70.5
    & 69.2
    & 65.9
    & 70.7
    & 64.5
    & 69.1
    & 62.7
    & 49.4
    & 57.4
    & \textbf{30.2}
    & 88.3
    & 61.8
    & 60.7
    & 48.8
    & 62.7 \\
      CoLA
    & 69.3
    & 68.6
    & 67.6
    & 63.1
    & 62.9
    & 59.3
    & 66.6
    & 61.3
    & 51.7
    & 50.2
    & 31.2
    & \textbf{81.0}
    & 58.5
    & 56.7
    & \textbf{45.3}
    & 59.6 \\
      \rowcolor[gray]{0.9}\textbf{Ours}
    & \textbf{68.5}
    & \textbf{67.1}
    & \textbf{66.3}
    & \textbf{62.6}
    & \textbf{62.8}
    & \textbf{57.4}
    & \textbf{65.9}
    & \textbf{60.0}
    & \textbf{51.2}
    & \textbf{49.8}
    & 31.3
    & 82.1
    & \textbf{56.0}
    & \textbf{55.4}
    & \textbf{45.3}
    & \textbf{58.8} \\
\bottomrule[0.3mm]
\end{tabular}
\vspace{-5mm}
\label{tab:quantization_results}
\end{table*}

\smallskip
\nbf{Ablation Study.}
Our method consists of three components: model merging, ZO-SGD, and pre-processing of the model weights. We conducted an ablation study to verify the effectiveness of each component. We also examined the effectiveness of each sub-component utilized in our preprocessing of the model weights (NWS, RR, ARS). The results using the ImageNet-C dataset and ViT-B/16 are shown in \cref{tab:ablation_study}. We found that model merging significantly improves the performance compared to No Adapt, demonstrating its usefulness. We succeeded in further improving the recognition performance by introducing ZO-SGD-based TTA. Regarding the effectiveness of each sub-component in our preprocessing, all of them succeeded in showing their effectiveness. The result that NWS improved the performance aligns with findings from previous research on model merging~\cite{TIES, Model-Breadcrumbs}. 
Among the sub-components, RR was found to contribute the most to performance. Although the impact of ARS on performance was smaller compared to the other two, it was confirmed to be consistently effective.

\smallskip
\nbf{Dimension Reduction Analysis.}
Our method incorporates a model merging strategy and model weights preprocessing to overcome the difficulties in high-dimensional optimization of ZOO. We examined whether these factors are effective in reducing the optimization dimensionality and enhancing ZOO. We used the CIFAR100-C dataset for this analysis. The results are shown in \cref{fig:num_param_analysis}. First, model merging succeeded in a dramatic reduction of the dimensionality, which significantly reduces the high-dimensionalility difficulty of ZOO. RR also contributed to reducing the dimensionality down to the rank out of the number of the merging models. Finally, ARS allowed us to reduce the dimensionality even further. We also examined how much these dimensionality reduction techniques enhanced the effectiveness of ZOO and found that while a naive application of ZOO yielded negligible performance improvements, we achieved larger improvements as the dimensionality was reduced by the proposed techniques.

\smallskip
\nbf{Memory Usage Analysis.}
We examined whether our method can actually reduce the memory usage compared to the conventional BP-based TTA methods (ROID and DeYO). To ensure fairness, the batch size was kept the same for all methods. The results are shown in \cref{fig:memory_usage}. ROID and DeYO required many times more memory usage than the normal inference. In contrast, CoLA and ours, which do not depend on BP, were executable with only a fraction of additional memory usage. These results demonstrate that our method is memory-efficient and more suitable for resource-limited devices than the conventional BP-based TTA methods.

\smallskip
\nbf{Applicability to Quantized Models.}
On resource-limited devices, model quantization techniques are often used to reduce the computational requirements. Since it is difficult to build the computational graph on the quantized models and calculate loss gradients, BP-based TTA methods are not applicable. In contrast, our proposed method does not depend on BP, making it effective for such models. Here, we experimentally verify the effectiveness of our proposed method on these quantized models using a dynamic-quantized ViT-B/16 and ImageNet-C dataset. As shown in \cref{tab:quantization_results}, we confirmed in practice that our method is effective even for quantized models, and compared our method with a stronger baseline, CoLA, our method achieved superior performance improvements.

\section{Conclusion}
\label{sec:conclusion}
In this paper, we proposed a novel test-time adaptation (TTA) method for resource-limited devices based on the cross-device collaborative TTA (CDC-TTA) framework. We associated zeroth-order optimization and model merging with the CDC-TTA framework and efficiently addressed two issues: avoiding the resource-intensive BP process by introducing the forward-only zeroth-order optimization (ZOO) and mitigating the high-dimensional difficulty in ZOO by reducing the number of optimization parameters based on the model merging of the models from multiple other devices. In addition, we proposed a unique preprocessing technique to further enhance the synergic combination of ZOO and model merging for TTA. We empirically demonstrated the superiority of our methods on prediction accuracy over the existing CDC-TTA method on common corruption benchmarks and style-transferred benchmarks, and confirmed the validity of our method for quantized models where gradient computation is unavailable. We hope our work will foster the research of TTA toward resource-limited scenarios.

\smallskip
\nbf{Limitations.}
One limitation of the proposed method is that its accuracy does not yet reach the level of BP-based methods. We assume that it is attributed to two main factors. First, the precision of gradient calculation is lower compared to BP. Second, because we restrict the number of parameters to overcome the high dimensionality difficulty of ZOO, the degrees of freedom for adaptation are inherently limited. A promising strategy to overcome the dilemma between memory capacity and predictive accuracy is a hybrid approach that partially incorporates BP. Our future research direction involves exploring methods to dynamically balance BP-updated components with those updated by our proposed method, tailored to the specific memory constraints of the device.

{\small
\bibliographystyle{ieeenat_fullname}
\bibliography{11_references}
}

\clearpage
\appendix
\section{Algorithm Procedures}\label{sec:algorithm_procedures}

Our method is separated into 1) server-side preprocessing and 2) device-side test-time adaptation. As described in \cref{sec:method}, these two procedures are executable independently. In this section, we provide the details of these procedures. We first use the model weights $\{\boldsymbol{\theta}^{(n)}_m\}^{M,N}_{m=0,n=1}$, the initial merging coefficients $\{\boldsymbol{w}^{(n)}\}^N_{n=1}$ in \cref{alg:preprocess}, and obtain the reduced merging coefficients $\{\boldsymbol{w}^{(n)}_r\}^N_{n=1}$ and the reduced task matrices $\{V^{(n)}_r\}^N_{n=1}$. The outputs of \cref{alg:preprocess} are then utilized in \cref{alg:zo_sgd} and the merging coefficients $\{\boldsymbol{w}^{(n)}_r\}^N_{n=1}$ are updated. As we can see, the procedures performed on the server side and device side are completely independent, and therefore, resource-intensive computations like singular value decomposition are fully handled by the computationally powerful server-side processing, causing no adverse effects whatsoever on the device side.

We describe the details of the random seed trick. Typically, in ZO-SGD, we need to prepare perturbation vectors of the same size as the parameters. Then, we perform inferences on positively and negatively perturbed models and estimate pseudo-gradients from those calculated losses. However, this procedure requires storing the perturbations, which necessitates memory usage equivalent to the model size and can potentially cause resource constraint problems. The ``random seed trick'' proposed by Hoge et al.~\cite{MeZO} sets the same random seed before generating the perturbations instead of storing the perturbations. Specifically, as described in lines 1, 8, and 16 of \cref{alg:zo_sgd}, we fix the random seed (e.g., \texttt{torch.random.seed(s)}) before sampling the perturbations. Then, the perturbations sampled in lines 3, 10, and 18 are ensured to be identical to each other. In this process, we only need to store the seed value $s$, thereby avoiding the memory usage problem associated with storing the perturbation.

\begin{algorithm}[t]
\caption{Preprocessing for merge coefficients}\label{alg:preprocess}
\begin{algorithmic}[1]
\Require Model weights $\{\boldsymbol{\theta}^{(n)}_m\}^{M,N}_{m=0,n=1}$, initial coefficients $\{\boldsymbol{w}^{(n)}\}^N_{n=1}$, parameters $k$, $\epsilon$, $r_\text{max}$.
\Ensure Merging coefficients $\{\boldsymbol{w}^{(n)}_r\}^N_{n=1}$, refined task matrices $\{V^{(n)}_r\}^N_{n=1}$.
\For{$n = 1 \cdots N$}
    \For{$m = 1 \cdots M$}
        \State $\boldsymbol{\tau}^{(n)}_m \leftarrow \boldsymbol{\theta}^{(n)}_{l_m} - \boldsymbol{\theta}^{(n)}_{l_0}$
        \State $\hat{\boldsymbol{\tau}}^n_m \leftarrow \text{top}_k(\boldsymbol{\tau}^n_m)$ \Comment{Remove insignificant values.}
    \EndFor
    \State Concatenate $\{\boldsymbol{\tau}^n_m\}^M_{m=1}$ to get $\hat{T}^n$.
    \State Get the rank size $r$ by Eq.~\eqref{eq:adaptive_rank_selection}.
    \State Apply SVD to $\hat{T}^{(n)}$ and get $U^{(n)}_r$, $\Sigma^{(n)}_r$, and $V^{(n)}_r$.
    \State $\boldsymbol{w}^{(n)}_r \leftarrow \boldsymbol{w}^{(n)} U^{(n)}_r \Sigma^{(n)}_r$
\EndFor
\end{algorithmic}
\end{algorithm}

\begin{algorithm}[t]
\caption{ZO-SGD for merge coefficients.}\label{alg:zo_sgd}
\begin{algorithmic}[1]
\Require Merge coefficients $\{\boldsymbol{w}^{(n)}_r\}^N_{n=1}$, task matrices $\{V^{(n)}_r\}^N_{n=1}$, frozen base model weights $\boldsymbol{\theta}_{f_0}$, learnable base model weights $\boldsymbol{\theta}_{l_0}$, random seed $s$, algorithm parameters $d$, $\eta$.
\Ensure Updated merging coefficients $\{\boldsymbol{w}^{(n)}_r\}^N_{n=1}$.
\State Set random seed to $s$.
\For{$n = 1 \cdots N$}
    \State Sample perturbation $\boldsymbol{z}^{(n)} \sim \mathcal{N}(0, I_{r^{(n)}})$.
    \State $\boldsymbol{w}^{(n)}_r \leftarrow \boldsymbol{w}^{(n)}_r + \delta \boldsymbol{z}^{(n)}$
    \State $\boldsymbol{\theta}^{(n)}_{l^+} \leftarrow \boldsymbol{\theta}^{(n)}_{l_0} + \boldsymbol{w}^{(n)}_r V^{(n)}_r$
\EndFor
\State $\mathcal{L}^+ \leftarrow \mathcal{L}\left(\boldsymbol{\theta}_{f_0}, \boldsymbol{\theta}_{l^+}, \mathcal{B}\right)$
\State Set random seed to $s$.
\For{$n = 1 \cdots N$}
    \State Sample perturbation $\boldsymbol{z}^{(n)} \sim \mathcal{N}(0, I_{r^{(n)}})$.
    \State $\boldsymbol{w}^{(n)}_r \leftarrow \boldsymbol{w}^{(n)}_r - 2\delta \boldsymbol{z}^{(n)}$
    \State $\boldsymbol{\theta}^{(n)}_{l^-} \leftarrow \boldsymbol{\theta}^{(n)}_{l_0} + \boldsymbol{w}^{(n)}_r V^{(n)}_r$
\EndFor
\State $\mathcal{L}^- \leftarrow \mathcal{L}\left(\boldsymbol{\theta}_{f_0}, \boldsymbol{\theta}_{l^-}, \mathcal{B}\right)$
\State $\Delta \leftarrow (\mathcal{L}^+ - \mathcal{L}^-)/2\delta$
\State Set random seed to $s$.
\For{$n = 1 \cdots N$}
    \State Sample perturbation $\boldsymbol{z}^{(n)} \sim \mathcal{N}(0, I_{r^{(n)}})$.
    \State $\boldsymbol{w}^{(n)}_r \leftarrow \boldsymbol{w}^{(n)}_r + \delta \boldsymbol{z}^{(n)} - \eta \Delta \boldsymbol{z}^{(n)}$
    \State $\boldsymbol{\theta}^{(n)}_l \leftarrow \boldsymbol{\theta}^{(n)}_{l_0} + \boldsymbol{w}^{(n)}_r V^{(n)}_r$
\EndFor

\end{algorithmic}
\end{algorithm}

\section{Implementation Details}\label{sec:implementation_details}

\begin{table}[t]
    \centering
    \scriptsize
    \caption{\textbf{Merge Model Training Setting.}}
    \begin{tabular}{wl{15mm}wc{10mm}wc{45mm}}
        \toprule[0.3mm]
        \multicolumn{3}{c}{ROID} \\
                             & Optimizer    &  Settings \\
        \midrule
        WideResNet-20        & Nesterov SGD & $\eta=\num{1e-3}$, $\mu=0.9$, $\lambda=0.0$, $bs=200$ \\
        ResNeXt-29           & Adam         & $\eta=\num{1e-3}$, $\beta=0.9$, $\lambda=0.0$, $bs=200$ \\
        ResNet-50            & Nesterov SGD & $\eta=\num{5e-4}$, $\mu=0.9$, $\lambda=0.0$, $bs=64$ \\
        ViT-B/16             & Nesterov SGD & $\eta=\num{1e-3}$, $\mu=0.9$, $\lambda=0.0$, $bs=64$ \\
        \midrule[0.3mm]
        \multicolumn{3}{c}{DeYO} \\
                             & Optimizer    &  Settings \\
        \midrule
        WideResNet-20        & Adam         & $\eta=\num{1e-3}$, $\beta=0.9$, $\lambda=0.0$, $bs=200$ \\
        ResNeXt-29           & Adam         & $\eta=\num{1e-3}$, $\beta=0.9$, $\lambda=0.0$, $bs=200$ \\
        ResNet-50            & Nesterov SGD & $\eta=\num{2.5e-4}$, $\mu=0.9$, $\lambda=0.0$, $bs=64$ \\
        ViT-B/16             & Nesterov SGD & $\eta=\num{2.5e-4}$, $\mu=0.9$, $\lambda=0.0$, $bs=64$ \\
        \bottomrule[0.3mm]
    \end{tabular}
    \label{tab:model_training_setting}
\end{table}

\begin{table}[t]
    \centering
    \scriptsize
    \caption{\textbf{Hyperparameter Settings.} We set the learning rate $\eta$, the perturbation size $\delta$, the filtering threshold $k$, the clamp size $c_\text{clamp}$, the maximum size of the matrix rank $r_\text{max}$, the matrix recovery error parameter $\epsilon$, and the batch-size $bs$ for each network architecture. We also set the size of initial merging coefficients $c_\text{init}$ for each benchmark setting.}
    \begin{tabular}{wl{22mm}wc{4mm}wc{4mm}wc{4mm}wc{4mm}wc{4mm}wc{4mm}wc{4mm}}
        \toprule[0.3mm]
                             & $\eta$         & $\delta$   & $c_\text{clamp}$ & $k$  & $r_\text{max}$ & $\epsilon$ & $bs$ \\
        \midrule
        WideResNet-20        & $\num{5e-3}$   & 1.0        & 5.0              & 40\% & 5              & \num{1e-2} & $200$\\
        ResNeXt-29           & $\num{5e-3}$   & 0.25       & 5.0              & 40\% & 5              & \num{1e-3} & $200$\\
        ResNet-50            & $\num{5e-3}$   & 0.25       & 5.0              & 40\% & 5              & \num{1e-2} & $64$\\
        ViT-B/16             & $\num{2.5e-3}$ & \num{1e-4} & 5.0              & 40\% & 5              & \num{1e-4} & $64$\\
        ViT-B/16 (Quantized) & $\num{2.5e-3}$ & \num{1e-1} & 5.0              & 40\% & 5              & \num{1e-4} & $64$\\
        \bottomrule[0.3mm]
    \end{tabular}
    \centering
    \begin{tabular}{wl{3mm}wc{17mm}wr{10mm}wc{3mm}wc{3mm}wc{3mm}wc{3mm}wc{3mm}}
        \toprule[0.3mm]
                        & Common Corruption &           & \multicolumn{5}{c}{Style Transferred Benchmarks} \\
                        & Benchmarks        & \#Domains & 10   & 20   & 30   & 40   & 50 \\
        \midrule
        $c_\text{init}$ & 0.10              &           & 0.35 & 0.15 & 0.10 & 0.10 & 0.05 \\
        \toprule[0.3mm]
    \end{tabular}
    \label{tab:hyperparameters}
\end{table}

In this work, we selected the affine parameters of the normalization layers (e.g., batch normalization layer normalization) as the trainable parameters $\boldsymbol{\theta}_{l}$, while the other model weights were treated as fixed parameters $\boldsymbol{\theta}_{f}$. When training the merge source models, $\boldsymbol{\theta}_l$ were optimized using existing TTA algorithms, DeYO~\cite{DeYO} and ROID~\cite{ROID}. The hyperparameters utilized in these trainings were set as \cref{tab:model_training_setting}. Note that the other hyperparameters were set to the default values of their respective algorithms.

In the device-side test time adaptation, we configured the initial merging coefficients to be uniform across all models. That is, we set the initial coefficients as $w^{(n)} = c_{\text{init}} \times \textbf{1}$, where $\textbf{1}$ denotes an all-ones vector. Since the value of $c_{\text{init}}$ needed to be adjusted according to the number of models to be merged (= \#Domains), we configured them for each benchmark. The other algorithm hyperparameters were configured for each network architecture. These hyperparameter settings are summarized in \cref{tab:hyperparameters}. We excluded the merging coefficients of the last block of WideResNet-20, ResNeXt-29, and ResNet-50, and those of the last three blocks of ViT-B/16, from the optimization target. We utilize the same loss function $\mathcal{L}$ as used in the merge source models.

\begin{figure}[t]
\begin{center}
\subfloat[][]{\includegraphics[width=.32\linewidth]{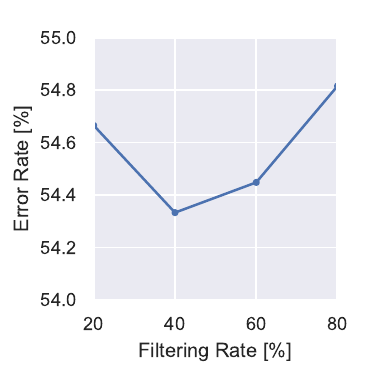}
\label{fig:filtering_rate}}
\hfill
\subfloat[][]{\includegraphics[width=.32\linewidth]{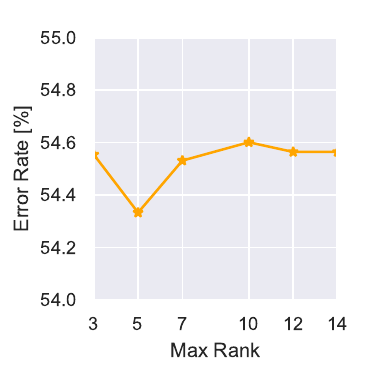}
\label{fig:max_rank}}
\hfill
\subfloat[][]{\includegraphics[width=.32\linewidth]{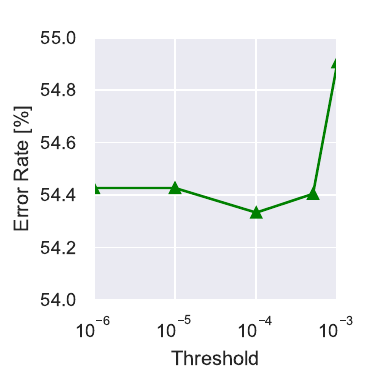}
\label{fig:matrix_threshold}}
\end{center}
\caption{{\bf Sensitivity to hyperparameters.} We analyzed (a) filtering ratio $k$, (b) max rank value $r_\text{max}$, and (c) matrix recovery error threshold $\epsilon$.}
\label{fig:hyperparameter_analysis}
\end{figure}

\smallskip
\nbf{Analysis.} We analyzed the setting of the three key hyperparameters introduced in our method. The results are shown in \cref{fig:hyperparameter_analysis}. 
In \cref{fig:filtering_rate}, we can see that setting the filtering rate $k$ to excessive or insufficient values results in degraded results, but as long as the value is set within a reasonably appropriate range, the performance difference within that range is not so significant.
In \cref{fig:max_rank}, the setting of maximum rank $r_\text{max}$ to $5$ is optimal, and under the other settings, the performance was degraded to some extent. We consider this is because a rank setting that is too small deteriorates the precision of the low-rank approximation, while a larger rank adversely affects the convergence of ZOO.
In \cref{fig:matrix_threshold}, the small matrix recovery error $\epsilon$ resulted in a slight increase in the error rate, but increasing $\epsilon$ caused relatively large performance drops. Setting $\epsilon$ to larger values leads to a reduction in the rank size, which results in the same configuration as setting a small $r_\text{max}$.

\end{document}